\documentclass[conference]{IEEEtran}
\IEEEoverridecommandlockouts
\usepackage{cite}
\usepackage{amsmath,amssymb,amsfonts}
\usepackage{algorithmic}
\usepackage{graphicx}
\usepackage{textcomp}
\usepackage{xcolor}
\usepackage{float}
\usepackage{graphicx}
\usepackage{subfigure}
\usepackage{algorithm}
\usepackage{threeparttable}
\usepackage{CJKutf8}

\def\BibTeX{{\rm B\kern-.05em{\sc i\kern-.025em b}\kern-.08em
    T\kern-.1667em\lower.7ex\hbox{E}\kern-.125emX}}
\begin{document}

\title{SmartRSD: An Intelligent Multimodal Approach to Real-Time Road Surface Detection for Safe Driving }

\author{\IEEEauthorblockN{Adnan Md Tayeb, Mst Ayesha Khatun, Mohtasin Golam, Md Facklasur Rahaman,\\ Ali Aouto, Oroceo Paul Angelo, Minseon Lee, Dong-Seong Kim, Jae-Min Lee, and Jung-Hyeon Kim
}

\IEEEauthorblockA{\textit{Networked Systems Laboratory, Department of IT Convergence Engineering,} \\
\textit{Kumoh National Institute of Technology, Gumi, South Korea.}\\
{(mdtayebadnan,20226138, golam248, facklasur, ali.aouto, oroceopaul, 20236117, dskim,ljmpaul, trizkim)@kumoh.ac.kr} \\}
}

\maketitle

\begin{abstract}
Precise and prompt identification of road surface conditions enables vehicles to adjust their actions, like changing speed or using specific traction control techniques, to lower the chance of accidents and potential danger to drivers and pedestrians. However, most of the existing methods for detecting road surfaces solely rely on visual data, which may be insufficient in certain situations, such as when the roads are covered by debris, in low light conditions, or in the presence of fog. Therefore, we introduce a multimodal approach for the automated detection of road surface conditions by integrating audio and images. The robustness of the proposed method is tested on a diverse dataset collected under various environmental conditions and road surface types. Through extensive evaluation, we demonstrate the effectiveness and reliability of our multimodal approach in accurately identifying road surface conditions in real-time scenarios. Our findings highlight the potential of integrating auditory and visual cues for enhancing road safety and minimizing accident risks.
\end{abstract}


\section{Introduction}
The real-time detection of road surface conditions is an essential measure to guarantee the safety of traffic, especially in the case of adverse weather conditions, including rain, snow, or ice. There is a direct link between tire-road friction, road surface condition, and the frequency of car crashes [1–3]. According to the recent statistics, accidents due to weather inflict almost 6,000 deaths and 445,000 injuries yearly. Wet weather stands out as the main cause in most of these incidents, with 73 percent of crashes happening on wet roads and 46 percent during rain. Additionally, over 200,000 accidents are associated with sleet and snow, with icy road conditions contributing to approximately 150,000 collisions [4].These statistics underscore the urgency of developing robust detection systems capable of identifying road surface condition in real-time. Furthermore, the impact of accurate surface condition detection extends beyond individual vehicles to encompass broader transportation networks. Timely and precise detection of road surface conditions enables transportation authorities to implement proactive measures, such as road maintenance and treatment strategies, to mitigate risks and ensure the smooth flow of traffic. The road condition detection is also important for the growing area of autonomous driving. Self-driving cars depend greatly on precise up-to-date information to quickly make decisions, particularly in challenging weather conditions when roads can quickly change. Therefore, the detection of road surface types has been a focal point of research for several decades, attracting considerable attention within academic and industry circles. Numerous ongoing projects in the field underscore the sustained interest and commitment to advancing our understanding and capabilities in road surface detection.

However, The majority of surface condition assessment techniques rely on visual data, which may be insufficient in certain situations, like low visibility, when the road is obstructed by debris, or in the presence of fog. Additionally, visual information can result in false information, particularly in varying lighting situations[9]. In low light, for example, a dry surface may appear wet, while surfaces reflecting light can give the impression of a dry surface being wet. Consequently, using only images may not suffice for accurate road condition detection.
In response to these challenges, there has been growing interest in integrating multiple modalities of sensory information to enhance the accuracy and reliability of road surface condition detection systems. Detecting road conditions automatically by utilizing both visual and audio inputs may play a vital role in enhancing driver assistance systems (EDAS) and ultimately enhancing driver safety in the future. Furthermore, autonomous and semi-autonomous vehicles must be conscious of the state of the roads to adjust the vehicle's settings automatically to maintain a safe distance or reduce speed before entering the curve to the car ahead.

\section{Methodology}

\begin{figure*}[t!]
    \centering
    \includegraphics[width=0.9\linewidth]{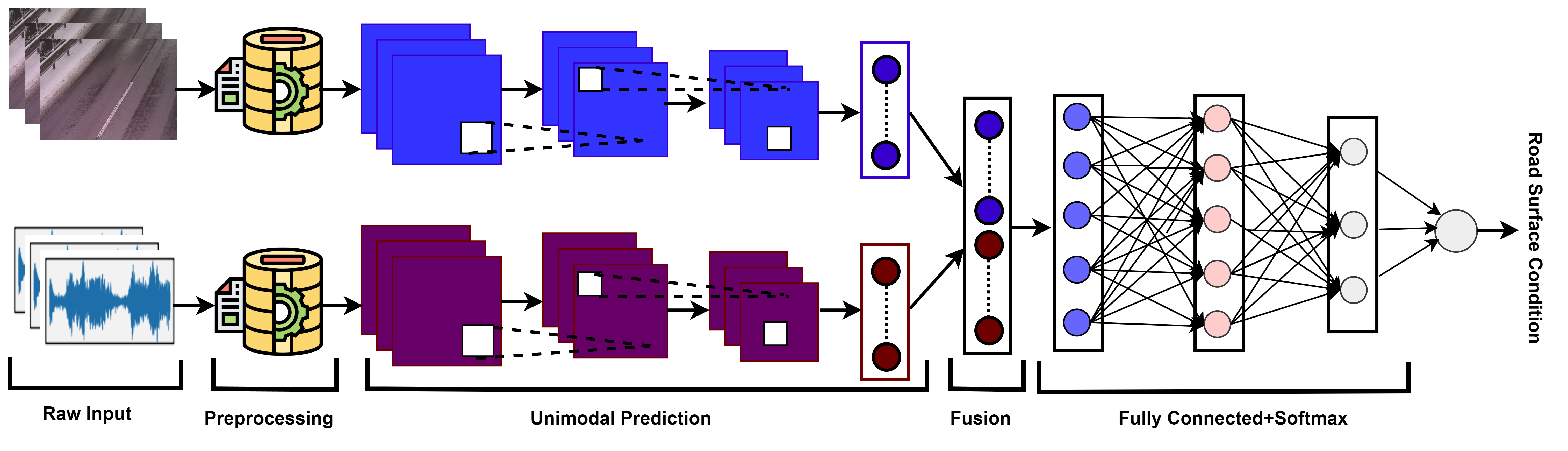}
    \caption{ Overview of the Proposed Multimodal Fusion}
    \label{fig:acdc_isic}
\end{figure*}

This section will delve into the methodology employed in this study, providing an overview of multimodal fusion techniques, the approach to data collection, and the preprocessing steps undertaken. Additionally, it will discuss the weighted fusion technique utilized in this study.

\subsection{Multimodal fusion techniques}\label{AA}
Multimodal fusion techniques involve combining data or information from different sources or methods to enhance the performance or comprehension of a system or application. Multimodal fusion is especially important in affect recognition tasks, where analysis relies on data from various sources like audio, visual, physiology, and more, showing significant value. Its use in classification tasks is significant, as combining data from different sources can enhance the feature space, strengthen the accuracy of models, and provide a more complete understanding of complex phenomena.
The literature describes three main joint fusion strategies: feature-level, decision-level (or score-level), and model-level fusion. Every approach has unique benefits that are customized to meet different use cases.

\begin{itemize}
    \item Feature-level fusion: Feature-level fusion is the process of combining features from various sources into one feature vector in order to incorporate information from all sources. This method reflects the way humans process information, combining various inputs like sound and sight before making forecasts. Feature-level fusion requires large datasets due to combining information from multiple modalities, but it offers the advantage of robustness to missing data from one modality, enabling continued predictions [7,8].
    \item Decision-level (Score-level) fusion: In decision-level fusion, each source is used separately to make predictions, and then the results from each source are combined. One downside is that if data from one source is missing, we can't make full use of it. Fusion methods can be straightforward, like taking a majority vote for classification tasks, or more complex, like learning weights. For regression tasks, we can train a linear regressor using predictions from each source and use its weights for fusion.
    \item Model-level (Hybrid-level) fusion: Model-level fusion combines the best parts of both feature-level and decision-level fusion methods. For example, it might mix features from some sources and then combine those predictions with scores from other sources that were looked at separately. This approach is adaptable and can adjust to what the task needs. An example is the method from , which mixes results from feature-level fusion with scores from separately processed sources. This hybrid method aims to get the advantages of mixing features from some sources while still considering the unique information from others. Using this fusion technique can make affect recognition tasks work better, especially when dealing with lots of different data from many sources.
\end{itemize}

\subsection{Dataset}
\begin{figure*}[t!]
    \centering
    \includegraphics[width=0.9\linewidth]{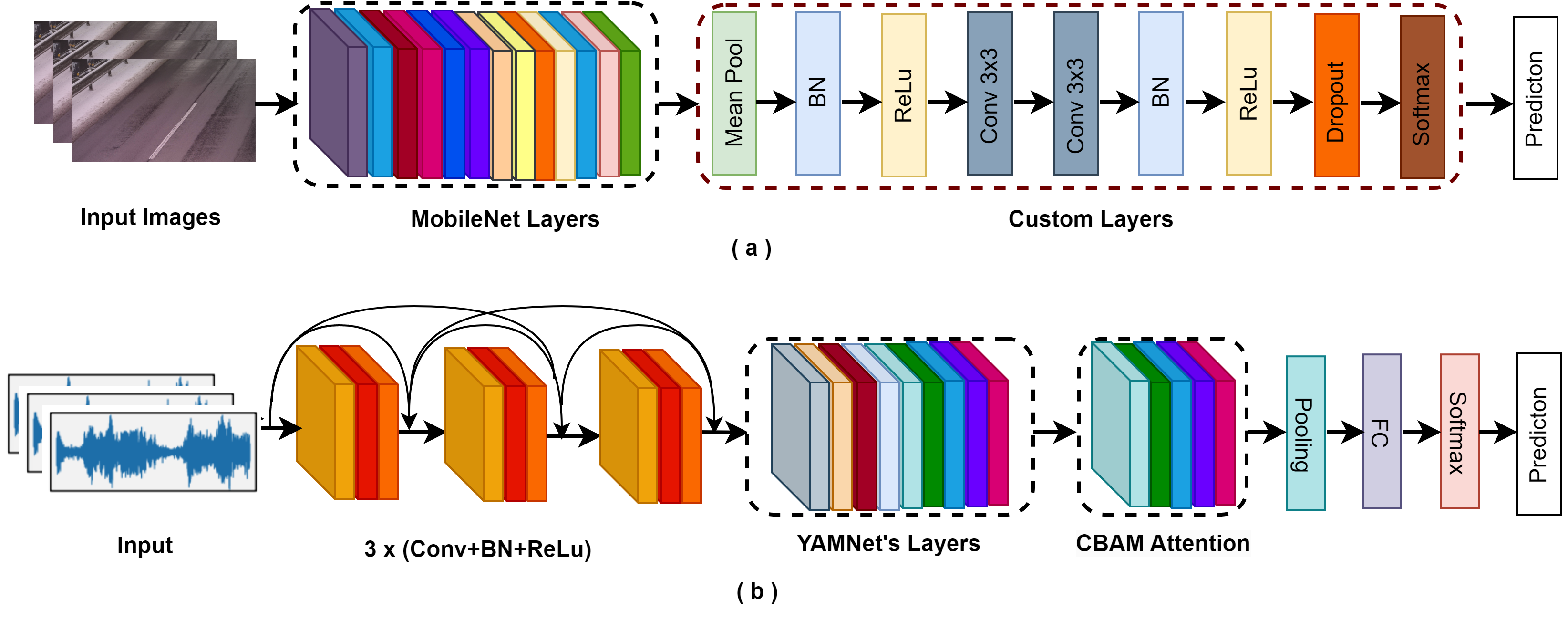}
    \caption{ Overview of the Proposed (a) Improved MobileNet, and (b) Improved YAMNet architecture for SmartRSD}
    \label{fig:acdc_isic}
\end{figure*}

\begin{figure}[H]
    \centering
    \includegraphics[width=.8\linewidth, height=0.15\textheight]{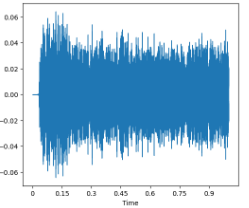}
    \caption{Raw Audio}
    \label{fig:enter-label}
\end{figure}

\begin{figure}[H]
    \centering
    \includegraphics[width=.8\linewidth]{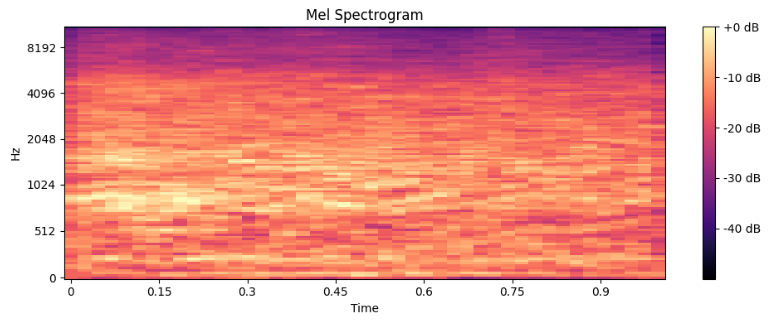}
    \caption{Spectogram}
    \label{fig:Spectogram}
\end{figure}

Data collection for this study involved setting up a testbed equipped with both a camera for capturing images of roads and a microphone for recording audio. Over the course of one year, observations were made and data was collected in various environmental conditions and seasons, including dry, rainy, and snowy. Various locations were used to gather a diverse range of road surface types and conditions for the images and audios obtained. This method enabled us to create a thorough collection of data that mirrors the situations vehicles face while driving.  We meticulously chose combinations of images from the collected data to represent the best and worst cases, covering all possible scenarios and ensuring the model's ability to learn from a variety of situations. Altogether, we have 8,580 images in our dataset, with 2,868 images showing dry conditions, 2,548 images showing wet conditions, and 2,354 images showing snow conditions. Furthermore, a total of 75 minutes of audio recordings were gathered and grouped into three categories: dry, wet, and snow.

\subsection{Data Preprocessing}
In our audio preprocessing pipeline, we standardized all audio clips by downsampling them to a uniform sampling rate of 16 kHz using the Sound library. A 1-second audio snippet is regarded as the input for all of our training and testing tasks, with each second of audio retaining the original recording level. To maintain consistency, we ensure that shorter audio clips are padded to match the 1-second duration. We generate Mel spectrograms (see Fig 3), a transformation that provides a detailed representation of the frequency composition of the audio signal over time. By converting the audio signal into a visual representation, the Mel spectrogram serves as the input for our machine-learning models. This approach enables us to leverage well-established image classification techniques, as the spectrogram essentially becomes an image representation of the audio data. By treating the audio data as images, we can apply methodologies and algorithms commonly used in image processing and classification tasks, thus enhancing our ability to analyze and interpret the audio signals effectively.

\subsection{Improved YAMNet architecture}
YAMNet utilizes MobileNetV1 as its base network and comes pre-trained on the Google AudioSet dataset, encompassing 521 audio events. Consequently, in our research, we train YAMNet on unbalanced data and achieve notable performance gains. Before the feature extraction phase, we perform resampling into 16,000 Hz with one-channel audio to ensure uniformity. YAMNet is a deep learning-based model, which means it automatically extracts audio features through its feature extraction layer. This layer extracts audio features in the form of spectrograms, which are then passed to improved MobileNet layers for classification.

This process allows us to leverage the power of deep learning to automatically extract relevant audio features, which are then utilized for accurate classification using the improved MobileNet layers. 

In our enhanced version of YAMNet (as shown in Fig 4, we've integrated three sets of deeply connected convolutional layers, batch normalization, and ReLU activations on top of the original YAMNet architecture. This addition has led to a notable improvement in performance. Moreover, the incorporation of deep connected skip pathways plays a crucial role in preserving information and ultimately enhancing classification accuracy. These skip pathways enable the model to retain important features while mitigating information loss, thereby contributing to more accurate and robust classification results. These enhancements elevate the capabilities of YAMNet by augmenting its capacity to capture intricate patterns and features within audio data, leading to improved classification accuracy and performance.

\subsection{Improved MobileNet}
MobileNet, developed by Google, is an open-source computer vision model designed for training classifiers. Depthwise convolutions employ one filter per channel, splitting the standard convolution into two separable layers: one for applying the filter and the other for concatenation. Additionally, pointwise convolutions of size $1\times1$ are used, merging their output with the depthwise convolution.

This approach offers a significant reduction in the number of parameters compared to traditional networks, resulting in a lightweight deep neural network architecture [5,6]. Given its lightweight nature and efficiency, we have chosen to employ this approach in our model, especially considering that our model will run in a mini PC environment.

To further enhance classification performance, we've incorporated additional layers into the architecture. As illustrated in Fig 5, these include Convolutional layers, pooling operations, and ReLU activations. Additionally, a dropout layer has been integrated to mitigate the risk of overfitting.

Our experimentation has demonstrated a substantial improvement in the performance of MobileNet with the inclusion of these custom layers. By leveraging the efficiency and effectiveness of the MobileNet architecture along with these additional layers, we've achieved enhanced classification accuracy while maintaining a lightweight and resource-efficient model suitable for deployment in resource-constrained environments such as mini PC setups.

\subsection{Weighted Fusion}
Given that the accuracy of image classification typically surpasses that of audio classification, combining results from both modalities may introduce a mismatch. To address this potential issue, we've implemented a weighted fusion technique, assigning slightly more weight to image predictions and less to audio predictions. Specifically, we've assigned a 60\%  weight to image predictions and 40\% to audio predictions initially and updated according to prediction accuracy.

\begin{algorithm}
    \caption{Weights assignment algorithm for Multimodal Classifier}
\label{alg:weights_assignment}

\begin{algorithmic}[1]
\REQUIRE Initial weights for $\alpha^0$ audio, and $I^0$ image, $N$ Number of epochs

\STATE procedure \textsc{Weights}($\alpha^0$, $I^0$, $N$)
\STATE \quad Train $\alpha^0$ with $\{W_{\alpha_i}\}^k_{i=1}$
\FOR {$i = 1, 2, ..., N$}
    \STATE \quad \quad process batch
    \STATE \quad \quad update weights $\alpha_i$ 
\ENDFOR
\STATE \quad Train $I^0$ with $\{W_{I_i}\}^k_{i=1}$
\FOR {$i = 1, 2, ..., N$}
    \STATE \quad \quad process batch
    \STATE \quad \quad update weights $I_i$ 
\ENDFOR
\STATE \quad Initialize $\omega^0 \leftarrow w1 \cdot \alpha^N + w2 \cdot I^N$ 
\STATE \quad Train Multimodal classifier with $\omega^0$
\FOR {$i = 1, 2, ..., N$}
    \STATE \quad \quad process batch
    \STATE \quad \quad compute fusion output
    \STATE \quad \quad compute loss
    \STATE \quad \quad update weights $\omega^0$ 
\ENDFOR
\STATE \quad return $\omega^0$ (final weights for the fusion model)
\STATE end procedure
\end{algorithmic}

\end{algorithm}

This weighted fusion approach ensures a balanced integration of information from both image and audio modalities, enhancing overall prediction accuracy while minimizing the risk of mismatch errors.

\section{Results and discussion}
In this study, we trained various combinations of unimodal models to evaluate prediction accuracy. We used MobileNet for image classification and YAMNet for audio classification as our base models. To enhance performance, we modified both models and tested the best possible combinations with our multimodal classifier to determine the highest accuracy.

\begin{table}[h]
\centering
\caption{Classification Results of the Proposed Improved(I) MobileNet and YAMNet with Multimodal Classifier}
\begin{tabular}{|l|c|c|c|c|}
\hline
\textbf{Model} & \textbf{Accuracy} & \textbf{Precision} & \textbf{Recall} & \textbf{F1-score} \\ \hline
MobileNet+YAMNet  & 0.9118 & 0.9111 & 0.9118 & 0.9114 \\ \hline
IMobileNet+YAMNet  & 0.9379 & 0.9241 & 0.9314 & 0.9254 \\ \hline
MobileNet+IYAMNet  & 0.9275 & 0.9341 & 0.9336 & 0.9354 \\ \hline
IMobileNet+IYAMNet  & 0.9491 & 0.93.87 & 0.9356 & 0.9384 \\ \hline
\end{tabular}
\label{tab:classification_results}
\end{table}

As Table \ref{tab:classification_results} illustrates, the best accuracy is achieved by the combination of the proposed Improved MobileNet and Improved YAMNet, reaching 94.91\%. This surpasses other combinations by approximately 2\%.
\section{Conclusion}
This research presents a multimodal approach that combines audio and images to detect road surface conditions in real-time. we aim to address the limitations of existing methods that solely rely on visual data, particularly in challenging environmental conditions such as low light, fog, or debris covered roads.

\section*{Acknowledgment}
This work was partly supported by Innovative Human Resource Development for Local Intellectualization program through the IITP grant funded by the Korea government(MSIT) (IITP-2024-2020-0-01612, 33.3\%) and by Priority Research Centers Program through the NRF funded by the MEST(2018R1A6A1A03024003, 33.3\%) and by the MSIT, Korea, under the ICAN support program(IITP-2024-00156394, 33.3\%) supervised by the IITP

\end{document}